\documentclass{article} 
\usepackage{iclr2027_conference,times}

\usepackage{amsmath,amsfonts,bm}

\def\eqref#1{equation~\ref{#1}}

\def\1{\bm{1}}

\DeclareMathAlphabet{\mathsfit}{\encodingdefault}{\sfdefault}{m}{sl}
\SetMathAlphabet{\mathsfit}{bold}{\encodingdefault}{\sfdefault}{bx}{n}

\usepackage{booktabs, makecell}
\usepackage{hyperref}
\usepackage{url}
\usepackage{paralist}
\usepackage{graphicx}
\usepackage{booktabs}
\usepackage[dvipsnames]{xcolor}
\usepackage{multirow}
\usepackage{array}
\usepackage{tcolorbox}
\usepackage{amsmath}
\usepackage{amssymb}
\usepackage{xcolor}
\usepackage{xspace}
\usepackage{tikz}
\usepackage{geometry} 
\usepackage{caption}  
\usepackage{enumitem}
\definecolor{promptrule}{HTML}{3C4A5A}
\definecolor{promptbg}{HTML}{F7F8FA}
\definecolor{promptsub}{HTML}{E4E8EE}
\definecolor{phcolor}{HTML}{B54708}
\usepackage{wrapfig}
\usepackage{pgfplots} 
\pgfplotsset{compat=1.18}
\usetikzlibrary{positioning,arrows.meta,shapes.geometric,fit,calc}
\usepackage[capitalise,noabbrev]{cleveref}
\usepackage[table]{xcolor}

\newcommand{\code}[1]{\texttt{\small #1}}

\newcommand{\projectname}{\textsc{MG-VQA}\xspace}

\newcommand{\tCount}{\code{Count}\xspace}
\newcommand{\tFind}{\code{Find}\xspace}
\newcommand{\tCompare}{\code{Compare}\xspace}

\newcommand{\tBeneath}{\code{Beneath}\xspace}

\definecolor{promptrule}{HTML}{3C4A5A}
\definecolor{promptbg}{HTML}{F7F8FA}
\definecolor{promptsub}{HTML}{E4E8EE}
\definecolor{phcolor}{HTML}{B54708}

\tcbuselibrary{skins,breakable}
 
\newcommand{\ph}[1]{{\color{phcolor}\textlangle\textit{#1}\textrangle}}
\newcommand{\tool}[1]{\texttt{\small #1}}
 
\tcbset{
  promptbox/.style={
    enhanced, breakable,
    colback=promptbg, colframe=promptrule,
    boxrule=0.5pt, arc=1.5pt,
    fonttitle=\small\bfseries, coltitle=white,
    boxsep=2pt, left=5pt, right=5pt, top=3pt, bottom=3pt,
    toptitle=2pt, bottomtitle=2pt,
    parbox=false,
  },
  promptsubtitle/.style={
    colback=promptsub, coltext=promptrule,
    boxrule=0pt, left=5pt, right=5pt, top=2pt, bottom=2pt,
  },
}
 
\newenvironment{toollist}
  {\begin{description}[leftmargin=0pt, labelsep=0.5em, itemsep=2pt,
                       topsep=3pt, parsep=0pt, font=\normalfont]}
  {\end{description}}

\title{ \projectname: Manipulation Grounded Visual Question Answering with VLMs}
\author{%
  \parbox[t]{\dimexpr\textwidth-2\tabcolsep\relax}{%
    \centering\bfseries
    \mbox{Vineet Bhat}\quad
    \mbox{Mikaela Angelina Uy}\quad
    \mbox{Siyi Chen}\quad
    \mbox{Chan Hee Song}\quad
    \mbox{Alex Zook}\quad
    \mbox{Jenai Xuning Yang}\quad
    \mbox{Stan Birchfield}\quad
    \mbox{Valts Blukis}\quad
    \mbox{Jonathan Tremblay}\endgraf
    \vspace{4pt}
    {\normalfont\small NVIDIA\endgraf}
  }%
}
\usepackage{textgreek}

\iclrfinalcopy 
\begin{document}

\maketitle

\begin{abstract} 

Vision-language models (VLMs) have shown promising spatial reasoning capabilities from static visual inputs, where the evidence needed to answer a question is available in the provided views. However, in cluttered environments, answer-relevant evidence may be occluded rather than absent: an object may lie beneath a pile, be covered by another object, or have identifying information facing away from the camera. Answering such questions requires physical interaction to reveal the hidden evidence. 
We formalize this setting as \emph{Manipulation-Grounded Visual Question Answering} (\projectname), where an agent answers a question about an initially cluttered scene by using manipulation as an intermediate evidence-gathering operation. We introduce \projectname-Bench, comprising 600 human-verified questions across four spatial reasoning tasks, and evaluate it in a cluttered tabletop interactive environment. 
Eight VLMs are evaluated under three levels of environment access: \emph{Direct} (image only), \emph{Perception} (pointing, segmentation, and scene graphs), and \emph{Manipulation} (perception, grasping, and pushing). 
Across all models, \emph{Direct} and \emph{Perception} achieve average success rates of 36.2\% and 37.6\%, respectively, only slightly above an image-blind chance baseline of 32.7\%. In contrast, \emph{Manipulation} raises average success to 56.8\% (40.8--83.0\% across models). 
Stronger tool-calling VLMs, GPT~6~Astra (83.0\%) and Gemini~3.8~Flash (69.3\%), search persistently and re-ground after unsuccessful interactions, while weaker models often answer without gathering sufficient evidence or stop prematurely after failed actions. Our results highlight the need for VLM agents that use manipulation for persistent, physically grounded evidence gathering and recovery. Code and benchmark is available here: \url{https://github.com/vineet2104/MG-VQA}.
\end{abstract}

\section{Introduction}
\label{sec:intro}

A household robot must answer questions about cluttered workspaces, such as 
``Is my charging cable on the table?'', 
``How many batteries are left in the pile?'', 
``What is underneath the blue bowl?'', or 
``Which labeled container is the largest?'' 
A static view may not reveal the evidence needed to answer: objects can be occluded, stacked, or oriented away from the camera. 
Answering reliably requires lifting, sliding, or repositioning objects to reveal hidden evidence.
Manipulation becomes a necessary information-gathering step whenever the relevant evidence is hidden.

\begin{figure}[ht]
      \includegraphics[width=\linewidth]{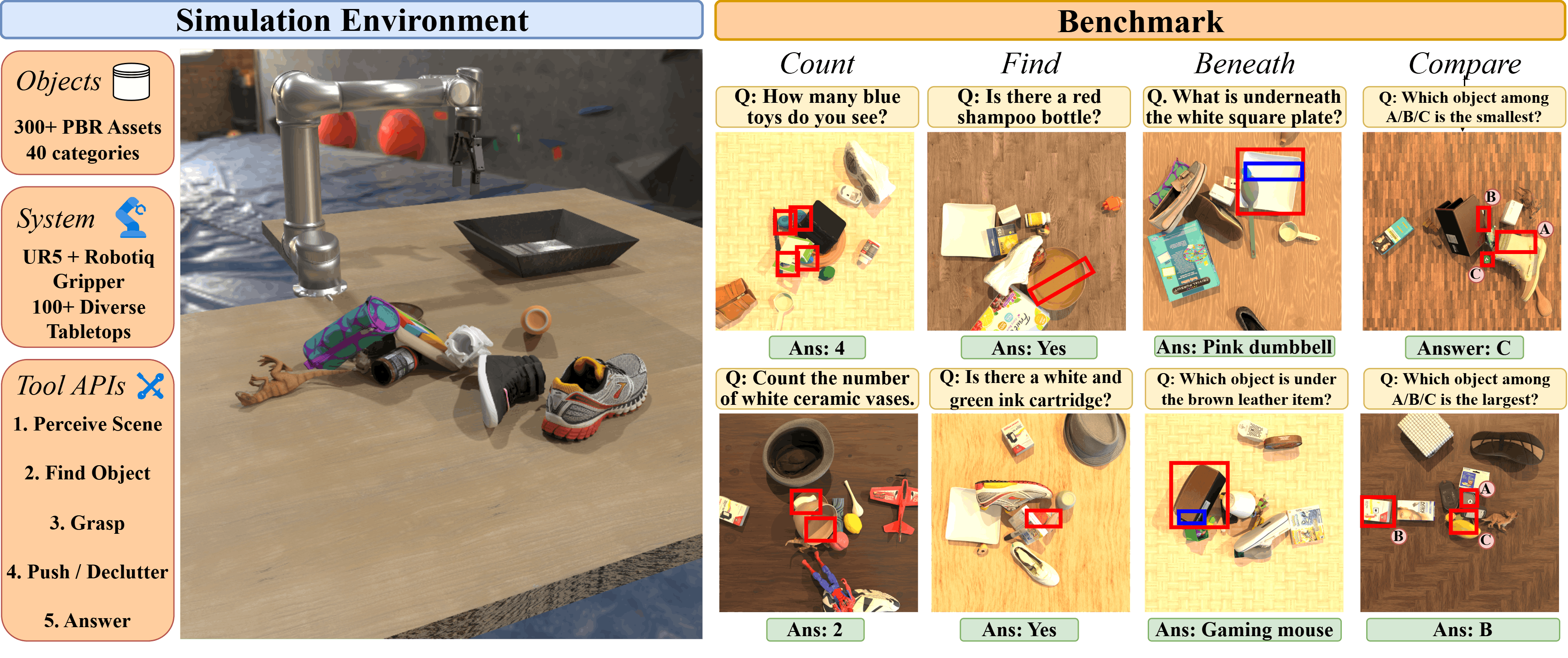}
      \caption{\projectname: a simulation environment and benchmark for manipulation-grounded spatial reasoning. Questions in \projectname-Bench run on a high-fidelity tabletop simulator (left) with a UR5 + Robotiq platform to physically interact with the scene. \projectname-Bench comprises 4 spatial reasoning tasks (right).}
      \label{fig:probe-bench-examples}
      \vspace{-15pt}
\end{figure}

Vision Language Models (VLMs) demonstrate broad spatial reasoning capabilities. They identify and refer to objects~\citep{you2024ferret}, describe spatial relations~\citep{chen2024spatialvlm}, estimate poses~\citep{bhat2025bop}, construct scene descriptions~\citep{pixel_to_graph}, and coordinate external tools~\citep{surismenon2023vipergpt}. 
VLMs are promising interfaces for robots operating in everyday environments. However, most evaluations of VLM spatial reasoning assume that the evidence needed to answer a question is available in static images, videos, or a fixed set of viewpoints. Such settings test whether a model can reason from available visual evidence, but not whether it can recognize when evidence is insufficient and physically uncover what is hidden.

This gap is particularly salient in dense tabletop clutter, where 
manipulation is not an oracle information request. A grasp can fail, an object can slip, and a push can disturb the scene without revealing the expected evidence. An effective agent 
must inspect the outcome of each manipulation action, determine whether its uncertainty has been resolved, and recover with an updated search strategy before committing to an answer. Current frontier VLMs have not been evaluated on this form of physically grounded evidence gathering.

We study this task as \emph{Manipulation-Grounded Visual Question Answering} (\projectname), where evidence needed to answer a question is occluded and must be revealed through manipulation (Figure~1). 
This distinguishes \projectname from conventional manipulation tasks, which aim at a desired final state of the environment, and from static visual question answering, which assumes the evidence is in the image. MG-VQA complements recent active and embodied spatial-reasoning works by focusing on cluttered tabletop question answering, where contact-based actions are physically consequential evidence-gathering operations.

To support this setting, we introduce \textbf{\projectname-Bench}, a human-verified benchmark of 600 questions over cluttered tabletop scenes covering four spatial reasoning tasks. Each question is constructed such that answer-critical evidence is occluded in the initial view. \projectname-Bench is evaluated in an interactive simulation environment \textbf{\projectname-Sim} with physically realized grasping and pushing actions executed by a UR5 robot. We evaluate frontier VLMs under three levels of access: \emph{Direct}, with only the initial image; \emph{Perception}, with additional scene-understanding tools; and \emph{Manipulation}, with grasping and pushing enabled. VLMs can access tool outputs such as 2D points, scene descriptions and segmentation masks, or invoke grasping and pushing primitives, before returning an answer to the question. This design separates models that reason only from visible evidence from those that use interaction to search for evidence, assess the outcome, and continue when necessary.

We find that \emph{Perception} tools provide limited gains over \emph{Direct} answering, while \emph{Manipulation} improves performance across all tasks. 
\projectname requires persistent, closed-loop evidence gathering and recovery, not merely spatial language understanding or access to tools.
Our contributions are:

{\setlength{\leftmargini}{2.0em}
\begin{compactenum}
    \item We formulate \textbf{Manipulation-Grounded Visual Question Answering (\projectname)}, where an agent answers questions about cluttered scenes using manipulation to reveal occluded, answer-relevant evidence. Manipulation is an intermediate information-gathering operation rather than the task objective.

    \item We develop and will release \textbf{MG-VQA-Sim}, an interactive evaluation environment for tool-calling VLMs that combines cluttered tabletop scenes, scene-understanding APIs, and physically realized grasping and pushing. Interactions carry physical consequences that persist, so agents must re-observe the scene and reason over a workspace their own actions have altered.

    \item We introduce \textbf{MG-VQA-Bench}, a human-verified benchmark of 600 questions across four spatial reasoning tasks: \textsc{Count}, \textsc{Find}, \textsc{Beneath}, and \textsc{Compare}. Questions are constructed so that answer-critical evidence is substantially occluded in the initial observation.

    \item We evaluate 8 VLMs with visual input only, with perception tools, and with manipulation tools. Physical interaction improves spatial reasoning, but models show persistent limitations: deciding when to search, gathering sufficient evidence, re-grounding after acting, and recovery from failed manipulation.
\end{compactenum}

\vspace{-10pt}

\section{Related Work}
\label{sec:related_work}

Robot manipulation benchmarks have used simulators to evaluate language-conditioned lifelong learning, long-horizon tabletop control, and embodied reasoning in GPU-parallelized simulation~\citep{NEURIPS2023_8c3c6668,mees2022calvin,taomaniskill3,mu2021maniskill,NEURIPS2021_021bbc7e,robocasa2024,Zhang_2025_ICCV,Mu_2025_CVPR,zhu2026toward,yang2026robolab,deng2020mqa}.
\projectname-Sim differs from these works in two ways: it is built for evaluating \emph{tool-calling VLMs} rather than VLA or RL policies, and each scene contains objects spawned in a tight, cluttered area to study recovery under challenging manipulation actions.
 Following similar motivations, ESI-Bench~\citep{hong2026esibench} evaluates embodied spatial reasoning through viewpoint changes in indoor environments, abstracting low-level execution to isolate information-seeking decisions.
%
 \projectname-Bench instead provides a benchmark for manipulation-grounded visual question answering, where VLMs must ground occluders, invoke grasp or push
 tools with contact-based outcomes unlike prior works that abstract away physics execution, and revise plans when interactions fail. \projectname-Bench combines real-scanned assets and
 600 human-validated questions over cluttered tabletop scenes. See \cref{app:extended-related-works} for an extended literature review.

\vspace{-5pt}

\section{\projectname: Manipulation-Grounded Visual Question Answering}
\label{sec:benchmark}

\projectname studies visual question answering in cluttered tabletop scenes where the evidence required to answer the question is not fully visible. 
An embodied agent cannot always answer a question from its initial observation: relevant objects, relations, or identifying information may be occluded by clutter. 
In such settings, the agent needs to physically alter the scene to reveal missing evidence 
and manipulation is an intermediate information-gathering step rather than the task objective.


This distinction makes the setting challenging. The agent does not receive an oracle observation after acting: a grasp may fail, a push may move several objects, and a partially successful interaction may leave relevant evidence occluded. To answer reliably, an agent must interpret the physical outcome of an action, and continue or revise its search when acquired evidence remains insufficient.


\vspace{-5pt}
\subsection{\projectname-Bench: Questions Requiring Manipulation-Grounded Evidence}
\label{subsec:tasks}

We instantiate the preceding motivation with \projectname-Bench, comprising four spatial reasoning tasks (150 questions each): \textsc{Count}, \textsc{Find}, \textsc{Beneath}, and \textsc{Compare}. Each task asks about a given scene, but places answer-critical evidence behind occlusion such that reliable answering requires targeted inspection, preventing an agent from answering from a single top-down observation:


\begin{itemize}[leftmargin=1.2em, itemsep=4pt, topsep=3pt, parsep=0pt]
    \item \textbf{\tCount: hidden inventory.}
    The agent is asked to count the number of instances of a target object present in the scene. Visible instances may understate the true count because additional instances can be covered or buried within a pile. The answer is an integer between 0 and 5.

    \item \textbf{\tFind: presence under occlusion.}
    The agent is asked if a named object is present in the scene. A ``yes'' answer requires localizing an instance with a 2D point. A target that is not visible may be present beneath clutter; therefore, an agent cannot infer absence if it cannot find the object in the initial image.

    \item \textbf{\tBeneath: revealing an occluded relation.}
    The agent is asked what lies underneath a referred object. The task requires moving the occluder, observing the exposed region, and identifying the revealed object. The agent chooses from four options, including ``None of the Above.''

    \item \textbf{\tCompare: inspecting hidden identifiers.}
    The agent is asked which of three objects marked with A/B/C labels is largest or smallest. Labels may not be initially visible, so the agent must uncover identifying evidence before associating each label with an object and comparing its physical size.
\end{itemize}

\noindent
Across all four tasks, manipulation serves a common purpose: it converts initially hidden scene information into observable evidence. However, the appropriate search strategy differs by task. \tCount{} and \tFind{} require the agent to inspect plausible occluders before concluding that no additional instance is present. \tBeneath{} requires targeted inspection of a queried cover object. \tCompare{} requires associating initially hidden identifiers with objects whose physical attributes must subsequently be compared. Example instances of all four task types are shown in \cref{fig:probe-bench-examples} and in \cref{app:benchmark_examples}. 

\paragraph{Construction principles.}
For \tCount{}, \tFind{}, and \tBeneath{}, we retain questions where answer-critical target object(s) have at most 10\% pixel visibility in the initial top-down view. For \tCompare{}, at least one of the A/B/C labels identifying the candidate objects is initially occluded. These criteria ensure that no question is answerable from the initial view alone.
\begin{wrapfigure}{r}{0.48\textwidth}
    \centering
    \vspace{-12pt}
    \includegraphics[width=\linewidth]{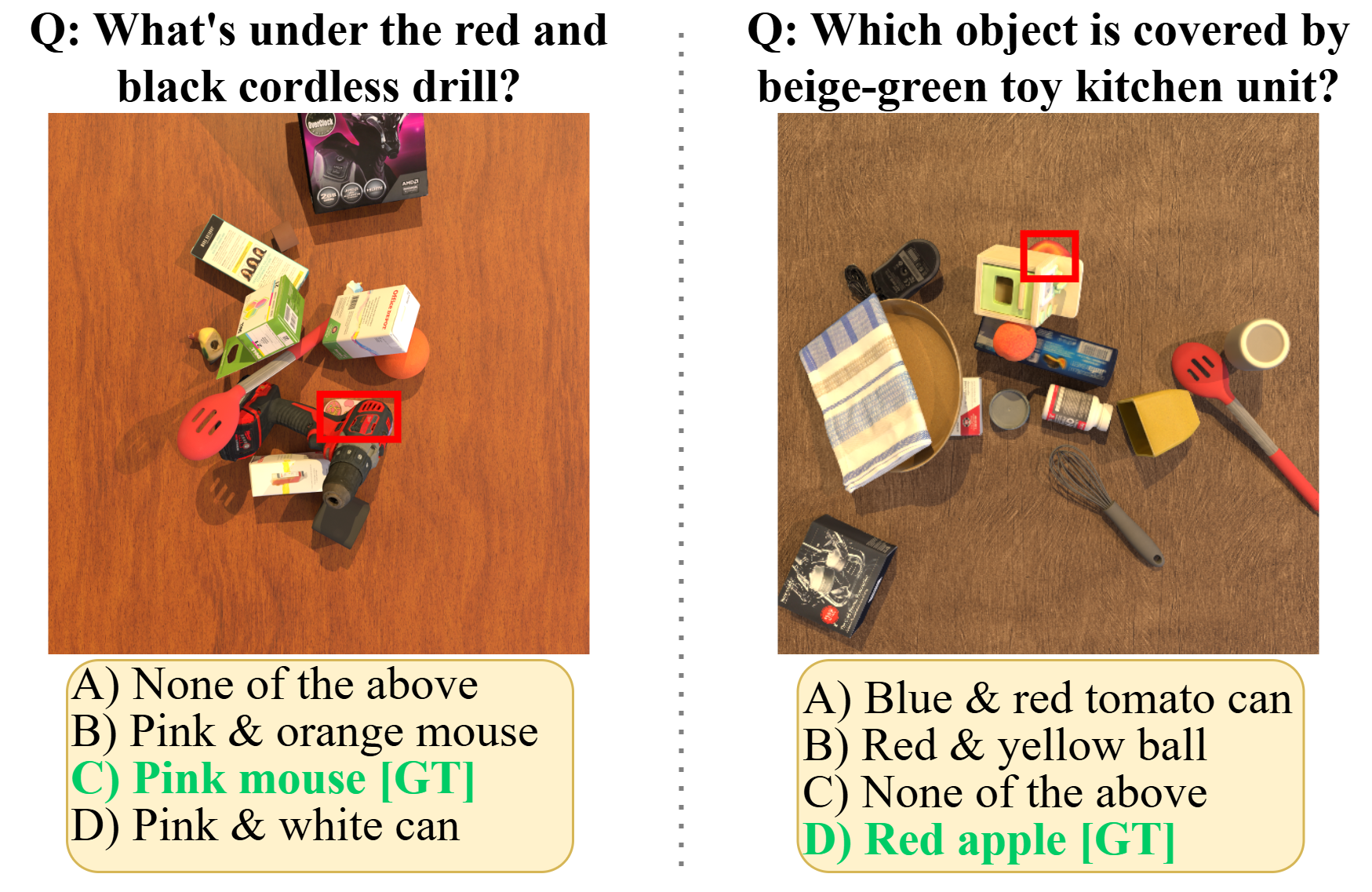}
    \caption{Examples where answer options are constructed to confound static question-answering.
    }
    \label{fig:example-ambig-options}
    \vspace{-25pt}
\end{wrapfigure}
We further construct a subset of instances with visually or semantically confounding distractors, illustrated in \cref{fig:example-ambig-options}. For example, when the hidden answer object is a red apple, the answer choices may also include another plausible object, such as a red and yellow ball. In these cases, a small visible patch of color or texture is insufficient to identify the hidden object; the agent must inspect the scene to obtain discriminative evidence. Such examples make it possible to test whether an agent genuinely gathers evidence through interaction rather than inferring an answer from a coarse visual cue.

\subsection{Benchmark Construction and Human Verification}
\label{subsec:construction}

\paragraph{Scene generation.}
We generate candidate scenes by sampling objects from an annotated pool of real-scanned everyday assets, placing up to 15 objects in a densely cluttered tabletop workspace, and simulating physics settling. For \tCount{}, we sample a queried object name and a target multiplicity from 0 to 5; when targets are present, they are placed before distractors such that they are likely to be occluded after settling. For \tFind{}, we generate equal numbers of present and absent cases, with present targets similarly placed underneath the clutter. For \tBeneath{}, we select cover--hidden-object pairs for which moving the cover substantially reveals the hidden object. For \tCompare{}, we select three objects with distinct physical volumes and apply standardized A/B/C labels to their object meshes on the visible side from the top view, with at least one label that is initially occluded. See \cref{app:scene_generation} for detailed scene-generation procedures.

After the scene settles, we compute ground-truth simulator state and retain only scenes satisfying the task-specific initial-occlusion criteria. VLMs receive only images and tool outputs, not privileged simulator state. We use Gemini~3~Flash to generate diverse question candidates, which are reviewed by human annotators.

\paragraph{Human verification.}
All candidate questions are reviewed by two human annotators using a custom verification interface. Annotators verify that: (i) question wording, candidate answers, and ground-truth answer are unambiguous; (ii) answer cannot be reliably determined from the initial observation alone; and (iii) the answer can be recovered after revealing the relevant evidence through feasible clutter removal. When a scene is valid but wording or answer options are ambiguous, annotators manually correct the metadata; otherwise, the instance is excluded. See \cref{app:human_verification} for details of the verification interface and protocol.

\subsection{\projectname-Sim: An Environment for Evidence Gathering and Recovery}
\label{subsec:probsim}

We develop \projectname-Sim as an interactive evaluation environment for VLMs answering questions in \projectname-Bench. The environment retains the physical consequences of information-gathering actions: a robot can clear clutter to expose hidden evidence, but actions may also fail, move unintended objects, or leave the answer only partially revealed. This allows us to evaluate whether a model can observe action outcomes and recover when its initial strategy is unsuccessful.

We implement \projectname-Sim using a tabletop environment, an attached UR5 robot with physically realized grasping and pushing, and render top-down RGB observations for VLMs. Each scene in \projectname-Bench contains up to 15 objects in a tightly bounded workspace and a clearing basket for temporarily relocating clutter. See \cref{app:simulation_stack} for the simulation and rendering stack, asset sources and annotations, workspace configuration, and scene-initialization procedure.

\paragraph{Tool interface.}
The agent interacts with the environment through four information-gathering tools and a terminal answer tool. The \code{perceive\_scene} tool returns a scene description, scene graph, visible-object list with 2D points, and pairwise spatial relations; \code{find\_object} returns a segmentation mask for a queried point corresponding to an object from the scene graph. The \code{grasp} and \code{push} tools provide complementary contact-based manipulation actions to uncover occluded evidence, while \code{answer\_question} terminates the episode and records the response. See \cref{app:toolshed} for full API specifications.

\paragraph{Physically consequential actions and recovery.}
Unlike oracle-based reveal operations, tool calls can have imperfect outcomes. A grasp can fail due to low confidence predicted grasps or object slippage, a push can produce partial displacement or collateral object motion, and an initially plausible action can leave the answer-critical evidence occluded. These outcomes are purposeful features of the environment: they require an agent to inspect whether an action had its intended informational effect, rather than assuming that an attempted grasp or push resolved the uncertainty.

For example, after a failed grasp, an agent may re-query the scene, identify a newly accessible region, push nearby clutter aside, and attempt a different grasp. Conversely, it may determine from the updated observation that it has already acquired enough evidence to answer. \projectname-Sim therefore supports evaluation of persistent, physically grounded search rather than one-shot action selection. We will release the environment, scene-generation pipeline, tool APIs, benchmark instances, and evaluation code upon publication.

\begin{table}[t]
\centering
\caption{
Success rates (\%) on 600 human-verified questions (150 questions per task type). \emph{Avg.} is the macro-average over the four tasks; \textbf{bold} marks the best result per column, \underline{underline} the second best.}
\label{tab:headline}

\footnotesize
\setlength{\tabcolsep}{2pt}
\renewcommand{\arraystretch}{0.95}
\begin{minipage}{0.9\linewidth}
\begin{minipage}[t]{0.48\linewidth}
\centering
\textbf{Open-source models}\par\smallskip
\resizebox{\linewidth}{!}{%
\begin{tabular}{@{}lccccc@{}}
\toprule
Mode & Count & Find & Beneath & Compare & Avg.\\
\midrule
Chance & 22.0 & 50.0 & 26.0 & 32.7 & 32.7 \\
\midrule
\multicolumn{6}{c}{\textit{Gemma~4~31B~IT}}\\
Direct & 24.0 & 38.0 & 33.3 & 35.3 & 32.7 \\
Perception & 22.7 & 46.7 & 28.7 & 43.3 & 35.3 \\
\rowcolor{gray!15} Manipulation & 32.0 & 50.7 & 52.7 & 42.0 & 44.3 \\
\midrule
\multicolumn{6}{c}{\textit{Kimi~K3}}\\
Direct & 24.7 & 39.3 & 29.3 & 52.7 & 36.5 \\
Perception & 25.3 & 44.7 & 32.0 & 60.7 & 40.7 \\
\rowcolor{gray!15} Manipulation & 40.0   & 53.3   & 74.7   & 58.7   & 56.7   \\
\midrule
\multicolumn{6}{c}{\textit{GLM~5.3}}\\
Direct & 24.7 & 38.7 & 23.3 & 34.0 & 30.2 \\
Perception & 21.3   & 42.0   & 27.3   & 39.3   & 32.5   \\
\rowcolor{gray!15} Manipulation & 25.3   & 47.3   & 46.7   & 44.0   & 40.8   \\
\midrule
\multicolumn{6}{c}{\textit{Qwen~3.8~Flash~Next}}\\
Direct & 28.7 & 40.0 & 34.0 & 52.0 & 38.7 \\
Perception & 24.0 & 38.0 & 31.3 & 47.3 & 35.2 \\
\rowcolor{gray!15} Manipulation & 36.7 & 42.0 & 66.0 & \underline{71.3} & 54.0 \\
\bottomrule
\end{tabular}%
}
\end{minipage}
\hfill
\begin{minipage}[t]{0.48\linewidth}
\centering
\textbf{Proprietary models}\par\smallskip
\resizebox{\linewidth}{!}{%
\begin{tabular}{@{}lccccc@{}}
\toprule
Mode & Count & Find & Beneath & Compare & Avg.\\
\midrule
Chance & 22.0 & 50.0 & 26.0 & 32.7 & 32.7 \\
\midrule
\multicolumn{6}{c}{\textit{Gemini~3.1~Pro}}\\
Direct & 24.7 & 47.3 & 42.7 & 48.0 & 40.7 \\
Perception & 27.3 & 48.7 & 36.0 & 50.0 & 40.5 \\
\rowcolor{gray!15} Manipulation & \underline{62.7} & \underline{66.7} & 66.7 & 62.0 & 64.5 \\
\midrule
\multicolumn{6}{c}{\textit{Gemini~3.8~Flash}}\\
Direct & 28.0 & 49.3 & 44.0 & 47.3 & 42.2 \\
Perception & 25.3 & 48.7 & 41.3 & 46.0 & 40.3 \\
\rowcolor{gray!15} Manipulation & 60.0 & 66.0 & \underline{80.0} & \underline{71.3} & \underline{69.3} \\
\midrule
\multicolumn{6}{c}{\textit{Claude~Opus~4.8}}\\
Direct & 23.3 & 38.7 & 28.0 & 35.3 & 31.3 \\
Perception & 22.7 & 45.3 & 27.3 & 41.3 & 34.2 \\
\rowcolor{gray!15} Manipulation & 25.3 & 47.3 & 51.3 & 42.0 & 41.5 \\
\midrule
\multicolumn{6}{c}{\textit{GPT~6~Astra}}\\
Direct & 33.3 & 31.3 & 33.3 & 50.0 & 37.0 \\
Perception & 32.7 & 44.0 & 36.0 & 57.3 & 42.5 \\
\rowcolor{gray!15} Manipulation & \textbf{87.3} & \textbf{80.7} & \textbf{88.0} & \textbf{76.0} & \textbf{83.0} \\
\bottomrule
\end{tabular}%
}
\end{minipage}
\end{minipage}
\vspace{-10pt}
\end{table}

\section{Results on \projectname-Bench}
\label{sec:eval}



We evaluate VLMs using no tools (\emph{Direct} mode), only perception tools \code{perceive\_scene} and \code{find\_object} (\emph{Perception} mode), and both perception and manipulation tools \code{grasp} and \code{push} (\emph{Manipulation} mode). We also include an image-blind \emph{Chance} baseline that answers ``0" for \tCount, ``No" for \tFind, ``None of the above" for \tBeneath and randomly chooses among A/B/C for \tCompare. \cref{tab:headline} shows results on \projectname-Bench, all values reported as success rate in $\%$.

Across all models, the use of manipulation tools improves success rate compared to the \emph{Direct} and \emph{Perception} modes.
Models differ in their gains in \emph{Manipulation} mode: GPT~6~Astra gains 40.5\%, while GLM 5.3, Opus 4.8, and Gemma 4 31B gain less than 10\% over their \emph{Perception} mode success rates.
%
Top-performing models manipulate more on average across all tasks (\cref{fig:manip-and-errors}, left).
\emph{Perception} mode, where VLMs use tool outputs like 2D points, segmentation masks and scene graphs leads to mixed results over \emph{Direct} mode across models, with a difference of -3.5\% to +5.5\% on average.
%
%
%
In \emph{Direct} and \emph{Perception} modes, average success rate across all models is limited to 25.8\% on \tCount, $33.0\%$ on \tBeneath, $46.2\%$ on \tCompare and $42.5\%$ on \tFind.
With \emph{Manipulation}, the models on average achieve 46.2\% (+20.4\%) on \tCount, 65.8\% (+32.8\%) on \tBeneath, 58.4\% (+12.2\%) on \tCompare and 56.8\% (+14.2\%) on \tFind.
Compared to the image-blind \emph{Chance} baseline, \emph{Direct} and \emph{Perception} mode results lie within 8\% on average success rate, underlining the need for physical scene interaction to answer questions in \projectname-Bench.

\begin{figure*}[t]
    \centering
    \includegraphics[width=\textwidth]{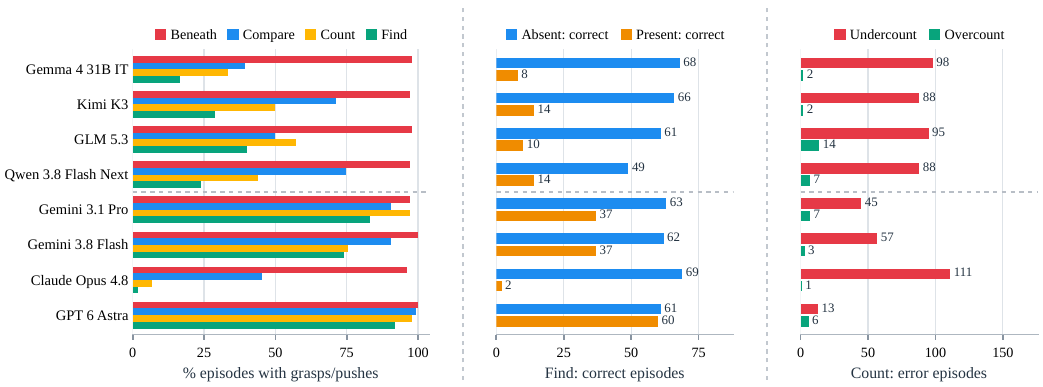}
    \caption{Manipulation and evidence gathering in \projectname-Bench. \emph{Left}: percentage of the 150 \emph{Manipulation}-mode episodes per model and task in which at least one \code{grasp} or \code{push} was used. \emph{Center}: correctly answered \tFind{} questions when the target is absent versus present, 75 episodes each. \emph{Right}: \tCount{} questions answered with an undercount versus an overcount; invalid count answers are excluded. Models are ordered as in \cref{tab:headline}; horizontal rule separates open-source from proprietary models. }
    \label{fig:manip-and-errors}
    \vspace{-15pt}
\end{figure*}

Models fail on \tFind and \tCount largely because they do not gather the evidence needed to answer. In the \tFind task, all models have better success rates in answering questions where the object is not present in the scene (\cref{fig:manip-and-errors}, center). In the \tCount task, model failures are dominated by undercounting rather than overcounting (\cref{fig:manip-and-errors}, right): VLMs may prematurely answer the question with the evidence of a few visible instances of a referred object, without investigating the rest of the scene for more instances. 
When VLMs manipulate objects, success rates rise sharply for object-present \tFind questions, from $7.3\%$ to $30.3\%$, and more modestly for object-absent questions, from $77.8\%$ to $82.8\%$.

\vspace{-10pt}

\subsection{Prompt ablations}
Models vary widely in their choice to invoke manipulation with substantial differences per task (\cref{fig:manip-and-errors}, left).
%
We introduce three prompt variations to study the behavior of VLMs under explicit manipulation instructions or auxiliary viewpoints: 
(i) \textit{Search} adds task-specific guidance at the beginning of the episode, encouraging manipulation of plausible occluders when objects are hidden; (ii) \textit{Search-persistent} repeats this guidance on every subsequent turn as a persistent message;
(iii) \textit{Multiview} adds an additional fixed, oblique RGB view along with the original top-down view for every model-turn. 
Table~\ref{tab:agentic-ablations}
shows that \textit{Search} guidance improves the average SR for both Kimi K3 (56.7\% to 61.8\%) and Opus 4.8 (41.5\% to 44.0\%).  \textit{Search-persistent} further improves scores for Kimi K3 (61.8\% to 66.0\%) but reduces it for Opus 4.8 (44.0\% to 41.5\%). 
Both these variants encourage improved interactions, as the number of episodes where these models interact with the scene increases (Kimi K3: 371 to 519 to 527, Opus 4.8: 225 to 293 to 305) along with the average manipulation tool calls per question. 

Explicit manipulation guidance can therefore encourage weaker models to manipulate more, but it only helps when the model can exploit what its actions reveal.
We also ran GPT~6 ~Astra with \textit{Search} ablation: the average success rate decreased from 83.0 to 80.5 while the number of episodes where manipulation occurred remained the same (584).
\textit{Multiview} ablation shows that an additional viewpoint improves performance without increasing manipulation actions. However, the improvement is not uniform across all tasks: SR for \tCount and \tBeneath increases (additional view may reveal hidden objects), but in \tCompare it is unchanged for Kimi and lower for Opus (badge labels A/B/C are only visible from top-down view). 

\begin{table*}[ht]
  \centering
  \small
  \vspace{-5pt}
    \caption{
    Ablations conducted on \projectname-Bench (600 questions); values reported in success rate (\%). Acted episodes contain at
  least one \code{grasp} / \code{push}. Ablations are separate additions.}
  \setlength{\tabcolsep}{3pt}
  \renewcommand{\arraystretch}{0.95}
  \fontsize{8.1}{9}\selectfont
  \begin{tabular}{llrrrrrrr}
    \toprule
    Model & Ablation & Beneath & Compare & Count & Find & Avg.
          & Acted /600 & Mean manip. /ep \\
    \midrule
    Kimi K3 & Baseline          & 74.7 & 58.7 & 40.0 & 53.3 & 56.7 & 371 & 1.7 \\
            & + Search         & 70.0 & 64.7 & 48.7 & 64.0 & 61.8 & 519 & 2.4 \\
            & + Search-persistent & 73.3 & 63.3 & 62.0 & 65.3 & 66.0 & 527 & 2.4 \\
            & + Multiview      & 76.0 & 58.7 & 54.0 & 60.7 & 62.3 & 385 & 1.6 \\
    \midrule
    Opus 4.8 & Baseline         & 51.3 & 42.0 & 25.3 & 47.3 & 41.5 & 225 & 1.0 \\
             & + Search        & 48.7 & 48.0 & 30.0 & 49.3 & 44.0 & 293 & 1.3 \\
             & + Search-persistent & 44.7 & 39.3 & 31.3 & 50.7 & 41.5 & 305 & 1.3 \\
             & + Multiview     & 54.7 & 37.3 & 34.0 & 48.7 & 43.7 & 215 & 0.7 \\
    \bottomrule
  \end{tabular}
  \vspace{-10pt}
  \label{tab:agentic-ablations}
\end{table*}

\subsection{What separates GPT~6~Astra from other VLMs}

\begin{wraptable}{R}{0.48\textwidth}
\centering
\vspace{-10pt}
\caption{
Median model turns / manipulation calls per question, computed over 150 questions per task.}
\vspace{-5pt}
\label{tab:search-effort-by-task}
\scriptsize
\setlength{\tabcolsep}{3pt}
\renewcommand{\arraystretch}{1.0}
\resizebox{\linewidth}{!}{%
\begin{tabular}{l c c c c}
\toprule
Model & \tCount & \tFind & \tBeneath & \tCompare \\
\midrule
Gemma~4~31B~IT      &  2 / 0   &  2 / 0 & 7 / 2 &  4 / 0   \\
Kimi~K3             &  3 / 0.5 &  2 / 0 & 7 / 2 &  7 / 2   \\
GLM~5.3             &  5 / 1   &  2 / 0 & 7 / 2 &  4 / 0.5 \\
Qwen~3.8~Flash~Next &  2 / 0   &  2 / 0 & 5 / 1 &  7 / 2   \\
Gemini~3.1~Pro      & 11 / 3   &  8 / 2 & 6 / 2 & 10 / 3   \\
Gemini~3.8~Flash    &  8 / 2   &  7 / 2 & 6 / 2 &  9 / 2   \\
Claude~Opus~4.8     &  2 / 0   &  2 / 0 & 6 / 2 &  5 / 0   \\
GPT~6~Astra         & 16 / 5   & 14 / 4 & 8 / 2 & 12 / 3   \\
\bottomrule
\end{tabular}%
}
\vspace{-5pt}
\end{wraptable}

GPT~6~Astra was released during the preparation of this paper and exhibits strong performance on all manipulation-grounded spatial reasoning tasks in \projectname-Bench. 
Its improvements can be attributed to: (1) better manipulation-driven search effort for occluded objects and (2) robust error recovery after an initial grasp failure.
Across all tasks, GPT~6~Astra executes more manipulation actions, averaging 3.9 per question, versus 0.9–2.9 for the other models.
Within each task, Astra uses more model turns (includes both perception and manipulation tools) than other models (\cref{tab:search-effort-by-task}).



GPT~6~Astra also shows better failure recovery patterns compared to other models (\cref{tab:recovery-flows}). 
After an initial grasp failure, every model chooses to push rather than retrying a grasp. 
The two lowest performing models, GLM 5.3 and Opus 4.8, directly choose to answer after a failed grasp in 46.3\% and 23.6\% of questions respectively.
Before invoking the \textit{push} tool, GPT~6~Astra prefers to generate a new scene graph followed by the segmentation tool (98.8\%). 
Failed grasps may move other scene objects, rendering the previous turn's scene graph stale. GPT~6~Astra consistently calls for a new scene graph before the next manipulation. We show a qualitative example in \cref{fig:astra-trajectory}.

\begin{table*}[ht]
\centering
\vspace{-5pt}
\caption{
Agent actions following failed grasps. $N$ denotes the number of episodes with at least one grasp failure. \emph{Next manipulation} groups these episodes by the next manipulation action: \code{push}, \code{grasp}, or none if the model answers directly. For episodes with a subsequent \code{push}, \emph{Re-grounding} shows the percentage in which the model updated the scene graph before segmentation (\emph{Scene + seg.}) or performed segmentation using an outdated scene graph (\emph{Seg.\ only}). Models are ordered as in \cref{tab:headline}.
}
\vspace{-5pt}
\label{tab:recovery-flows}
\small
\renewcommand{\arraystretch}{1.0}
\scalebox{0.9}{%
\begin{tabular*}{\textwidth}{@{\extracolsep{\fill}} l r r r r r r @{}}
\toprule
 & & \multicolumn{3}{c}{Next manipulation (\% of $N$)} & \multicolumn{2}{c}{Re-grounding (\% of pushes)} \\
\cmidrule(lr){3-5} \cmidrule(lr){6-7}
Model & $N$ & Push & Retry grasp & None & Scene\,+\,seg. & Seg.\ only \\
\midrule
Gemma~4~31B~IT      & 213 & 76.1 & 11.3 & 12.7 & 32.1 & 64.2 \\
Kimi~K3             & 246 & 85.8 &  7.7 &  6.5 & 43.1 & 55.5 \\
GLM~5.3             & 246 & 50.8 &  2.8 & 46.3 & 57.6 & 35.2 \\
Qwen~3.8~Flash~Next & 129 & 86.0 &  1.6 & 12.4 & 61.3 & 32.4 \\
Gemini~3.1~Pro      & 454 & 79.3 & 13.2 &  7.5 & 48.9 & 48.1 \\
Gemini~3.8~Flash    & 422 & 70.1 & 17.5 & 12.3 & 10.5 & 84.5 \\
Claude~Opus~4.8     & 182 & 48.4 & 28.0 & 23.6 &  4.5 & 88.6 \\
GPT~6~Astra         & 502 & 82.7 & 12.4 &  5.0 & 98.8 &  0.0 \\
\bottomrule
\end{tabular*}%
}
\vspace{-5pt}
\end{table*}

\begin{figure*}[ht]
  \centering
  \includegraphics[width=\textwidth]{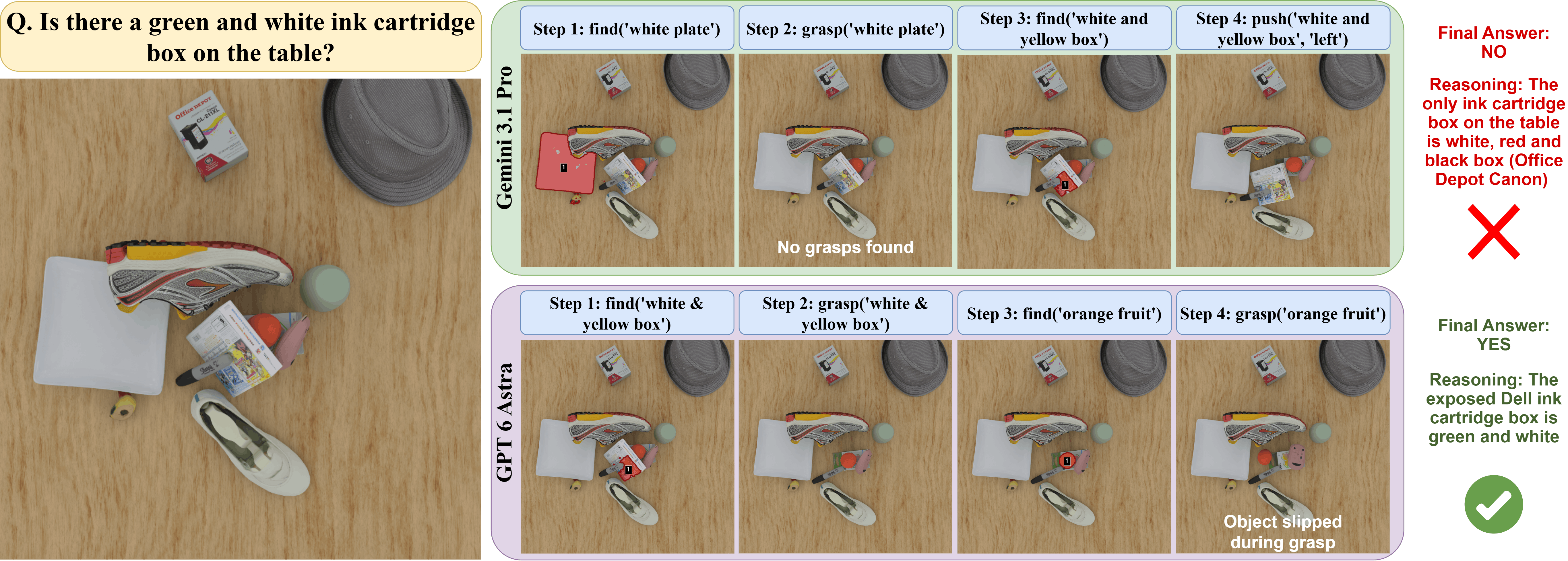}
  \caption{
  Gemini~3.1~Pro (top) attempts to grasp a flat plate, then displaces a different object with a push and concludes from the still-occluded scene that the target is absent. GPT~6~Astra (bottom) investigates the central pile of objects sequentially, answering only once the target is exposed. }
  \label{fig:astra-trajectory}
  \vspace{-10pt}
\end{figure*}

\subsection{Astra-assisted \projectname-Hard benchmark}
\label{sec:hard-scenes}

We utilize GPT~6~Astra as a coding agent to create a new scene generation algorithm for complex questions. The agent first analyzed the tool-calling behaviors across all models and was tasked with generating scenes that are likely to be challenging for existing models. Instead of the sequential clutter-spawning strategy used to create \projectname-Bench, the agent was given tools to spawn objects in specific locations, choose table textures and pick from an extended library of 50 objects (\textit{e.g.,} 3D objects that are not yet publicly available). 
The resulting scenes were manually verified and we will release it separately as the \projectname-Hard benchmark.
 %
\projectname-Hard comprises 40 questions, with 10 per task. We show example scenes in \cref{app:hard_scenes}. We evaluate 2 representative open-source models and 3 proprietary models on this subset, with results shown in \cref{tab:results_hard_questions}. 
All evaluated models underperform compared to \projectname-Bench, reporting 17.5\% -- 27.5\% on average success rate. \projectname-Hard breaks the correlation between success rate and manipulation actions, as Gemini 3.1 Pro with 3.75 average manipulation actions per question performs at the same level as Qwen 3.8 Flash Next with 0.8 manipulation actions per question. 

\vspace{-10pt}

\paragraph{Key findings.} GPT~6~Astra as a coding assistant exploits two vulnerabilities in current models. When the scene shows well-separated objects that appear to lie flat on the table, VLMs assume the isolated object arrangement does not contain any clutter. When smaller items are hidden underneath such objects, Kimi reasons ``flat boxes/DVD cases cannot hide cubes underneath in any meaningful way,'' while Gemini 3.8 Flash describes ``packages resting directly on
the table with no occlusions.'' Secondly, all models struggle in estimating the height of the objects in top-down view: six of GPT~6~Astra's eight comparison errors occur when it repeatedly judges a horizontally elongated object (e.g. ``butter box") larger than a vertically situated taller object (e.g. ``shampoo bottle") highlighting the challenges in 3D understanding in frontier VLMs.

\begin{table}[ht]
\centering
\vspace{-5pt}
\caption{{Success rates and manipulation statistics on the \projectname-Hard benchmark.} Results cover 40 questions, with 10 per task type. Success rates are reported as percentages. Column definitions follow \cref{tab:agentic-ablations}. \textbf{Bold} indicates the best result in each column, and \underline{underline} indicates the second best.}
\label{tab:results_hard_questions}
\vspace{2pt}
\footnotesize
\setlength{\tabcolsep}{3pt}
\renewcommand{\arraystretch}{0.95}
\scalebox{0.9}{%
\begin{tabular}{l c c c c c r r}
\toprule
Model & \tCount & \tFind & \tBeneath & \tCompare & Avg. & Acted\,/40 & Mean manip.\,/ep. \\
\midrule
\multicolumn{8}{c}{\textit{Open-source models}}\\
\addlinespace[1pt]
Kimi~K3             & 0.0              & \underline{10.0} & 10.0             & \underline{50.0} & 17.5             & 18 & 1.125 \\
Qwen~3.8~Flash~Next & 0.0              & 0.0              & \textbf{50.0}    & \textbf{60.0}    & \textbf{27.5}    & 17 & 0.800 \\
\midrule
\multicolumn{8}{c}{\textit{Proprietary models}}\\
\addlinespace[1pt]
Gemini~3.1~Pro      & \underline{10.0} & \textbf{20.0}    & 30.0             & \underline{50.0} & \textbf{27.5}    & \textbf{38} & \textbf{3.750} \\
Gemini~3.8~Flash    & 0.0              & \underline{10.0} & \underline{40.0} & \underline{50.0} & \underline{25.0} & 24 & 1.575 \\
GPT~6~Astra         & \textbf{20.0}    & \textbf{20.0}    & \underline{40.0} & 20.0             & \underline{25.0} & \underline{34} & \underline{3.450} \\
\bottomrule
\end{tabular}%
}
\vspace{-15pt}
\end{table}

\section{Conclusion}
\label{sec:conclusion}



Cluttered tabletop scenes challenge VLMs even with pointing, segmentation, and scene graphs, as answering spatial questions often requires manipulating objects to reveal hidden evidence. \projectname-Bench and \projectname-Hard evaluate this ability across frontier VLMs. Proprietary models outperform open-source models when manipulation is available (64.6\% vs. 49.0\%), with weaker models often reluctant to search through manipulation. The \projectname-Hard benchmark exposes additional biases, including inferring the absence of hidden objects from the spacing between visible ones. These findings highlight challenges in gathering and verifying evidence through interaction. We will release the benchmarks and code to support reproducible evaluation and progress in manipulation-grounded visual question answering.


\paragraph{Limitations.}
Our tool APIs and simulator address table-top environments with a fixed manipulator; extension to other tools, embodiments, and environments is left to future work.
VLMs in \projectname-Bench operate over a discrete fixed tool library, limiting their ability to recover from failures by creating new tools; future work will explore how VLMs can be given access to design better tools to manipulate objects.

\paragraph{AI use statement.}
We used generative AI tools as a coding assistant during the development of \projectname-Sim, the evaluation pipeline, plotting code for the figures in this paper, and to check the grammar and terminological consistency of the manuscript. We did not use generative AI tools to generate research ideas, to design \projectname-Bench or its evaluation protocol, or to conduct the literature review or select citations. Two uses fall within benchmark construction: object name annotations were produced by a VLM and subsequently verified by human annotators, and \projectname-Hard was constructed with AI coding assistance. All AI-assisted code was reviewed and verified by the authors through testing and inspection of its outputs, and all figures were checked against the underlying results. We take responsibility for the final content of this work, including all text, claims, and artifacts produced with the aid of generative AI.

\bibliography{iclr2027_conference}
\bibliographystyle{iclr2027_conference}

\appendix
\appendix
\crefalias{section}{appendix}
\crefalias{subsection}{appendix}
\crefalias{subsubsection}{appendix}

\providecommand{\supproot}{.}
\providecommand{\projectname}{MG-VQA}
\renewcommand{\projectname}{MG-VQA}
\providecommand{\code}[1]{\texttt{#1}}
\providecommand{\tCount}{\textsc{Count}}
\providecommand{\tFind}{\textsc{Find}}
\providecommand{\tBeneath}{\textsc{Beneath}}
\providecommand{\tCompare}{\textsc{Compare}}
\renewcommand{\thesection}{\Alph{section}}
\renewcommand{\thesubsection}{\thesection.\arabic{subsection}}
\newsavebox{\ExampleText}
\newcommand{\ExampleCell}[3]{%
  \begingroup
  \sbox{\ExampleText}{\begin{minipage}{0.190\textwidth}\raggedright
    \fontsize{6.2}{6.8}\selectfont #2\par\vspace{2pt}#3\par
  \end{minipage}}%
  \ifdim\dimexpr\ht\ExampleText+\dp\ExampleText\relax>51pt
    \PackageError{examples}{Question text exceeds cell}{Adjust example layout.}%
  \fi
  \begin{minipage}[t][148pt][t]{0.190\textwidth}\centering
    \includegraphics[width=92pt,height=92pt]{\supproot/#1}\par\vspace{3pt}
    \usebox{\ExampleText}\par
  \end{minipage}\endgroup
}

\section{Extended Related Work}
\label{app:extended-related-works}

\paragraph{Simulation environments for VLM-driven robotics.}
Robotic manipulation benchmarks have used simulators to evaluate language-conditioned lifelong learning~\citep{NEURIPS2023_8c3c6668}, long-horizon tabletop control~\citep{mees2022calvin}, and embodied reasoning in GPU-parallelized simulation~\citep{taomaniskill3,mu2021maniskill}.
They span household rearrangement~\citep{NEURIPS2021_021bbc7e}, generative kitchen environments~\citep{robocasa2024}, implicit-intent VLA tasks~\citep{Zhang_2025_ICCV}, dual-arm digital twins~\citep{Mu_2025_CVPR}, material-aware high-fidelity manipulation~\citep{zhu2026toward}, and broad competency axes across visual, procedural, and relational tasks~\citep{yang2026robolab}.
\projectname-Sim differs from these works in two ways: it is built for evaluating \emph{tool-calling VLMs} rather than VLA or RL policies, and each scene is designed with objects spawned in a tight, cluttered area to study recovery under challenging manipulation actions.

\paragraph{Embodied question answering.}
Embodied QA benchmarks progressed from navigation-based visual evidence gathering in synthetic scenes~\citep{embodiedqa} to real-world episodic-memory and active-exploration settings~\citep{OpenEQA2023}.
MQA~\citep{deng2020mqa} added object interaction in a V-REP bin environment and AI2-THOR with remote navigation~\citep{remqa,Kolve2017AI2THORAn}.
However, these benchmarks largely emphasize navigation and observation rather than physical manipulation.
ESI-Bench~\citep{hong2026esibench} evaluates embodied spatial reasoning through viewpoint changes and task-specific interactions in indoor environments, abstracting low-level execution to isolate information-seeking decisions. \projectname-Bench instead focuses on manipulation-driven evidence acquisition in dense tabletop clutter: VLMs must ground candidate occluders in images and invoke grasp or push tools whose outcomes depend on contact-based execution. This retains grounding errors, failed pickups, slippage, and collateral object motion within the reasoning loop, testing how VLMs revise their plans when an intended interaction does not produce the expected evidence.
Prior systems typically relied on disjoint modules for planning, VQA and manipulation, while recent tool-using VLMs enable robot agents that can reason, plan, and act to answer questions beyond static RGB scenes.
\projectname-Bench provides the first VLM benchmark to measure manipulation-grounded visual question answering in VLMs, combining perception and physically realized manipulation tools, real-scanned assets, and human-validated $600$-questions over cluttered tabletop scenes.


\paragraph{Spatial Reasoning and VLM Agents.} 
VLM-based robot perception has progressed from predicting language-conditioned spatial-affordance keypoints~\citep{yuan2024robopoint} to 3D-aware spatial referring and multi-step reasoning~\citep{zhou2025roboreferspatialreferringreasoning}, yet frontier VLMs still struggle with fine-grained interaction queries such as grasp and waypoint prediction~\citep{bhat2025bop}.
Work on LLM/VLM tool use has evolved from high-level action primitives for robot planning~\citep{singh2023progprompt,saycan2022arxiv} to code-as-policy agents over perception-control primitives~\citep{fu2026cap} and VLM agents that coordinate external perception tools~\citep{chen2026spacetools}.
These approaches support our central hypothesis that grounded interaction improves VLM-driven manipulation, but existing evaluations remain largely limited to navigation or static RGB-D scenes; \projectname-Bench instead studies cluttered tabletop settings where VLMs must decide whether and when to physically interact with the world through manipulation.

\section{\projectname-Bench Examples}
\label{app:benchmark_examples}\label{app:benchmark_catalog}

\begin{figure}[!htp]
\centering
\includegraphics[width=\textwidth,height=0.78\textheight,keepaspectratio]{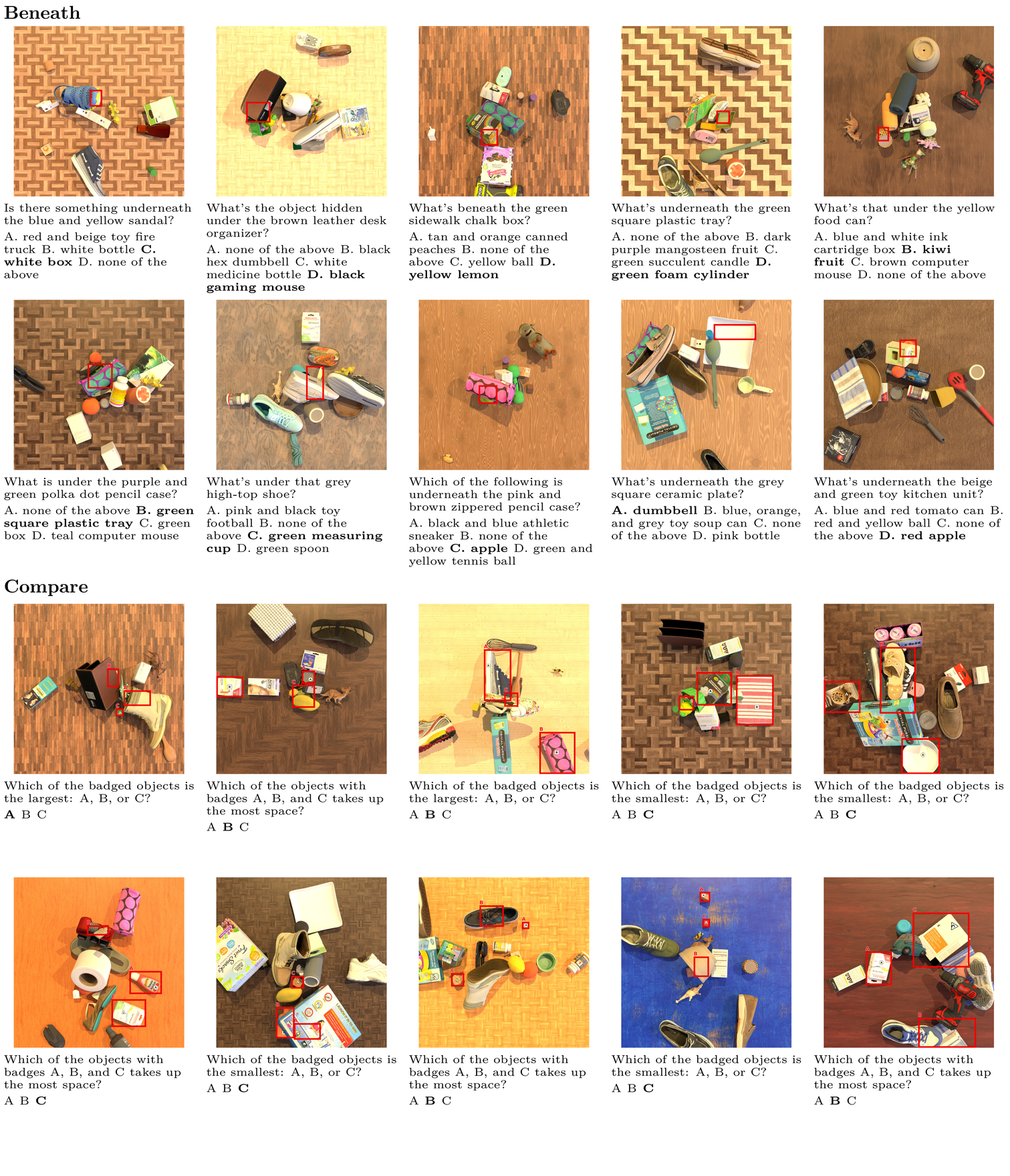}
\caption{\textbf{\projectname-Bench: Beneath and Compare.} Ten examples per task.
Correct answers are bold. Red boxes show privileged ground-truth target
extents, including hidden portions; these overlays are not agent inputs.}
\label{fig:appendix_benchmark_examples}
\end{figure}

\begin{figure}[!htp]
\centering
\includegraphics[width=\textwidth,height=0.78\textheight,keepaspectratio]{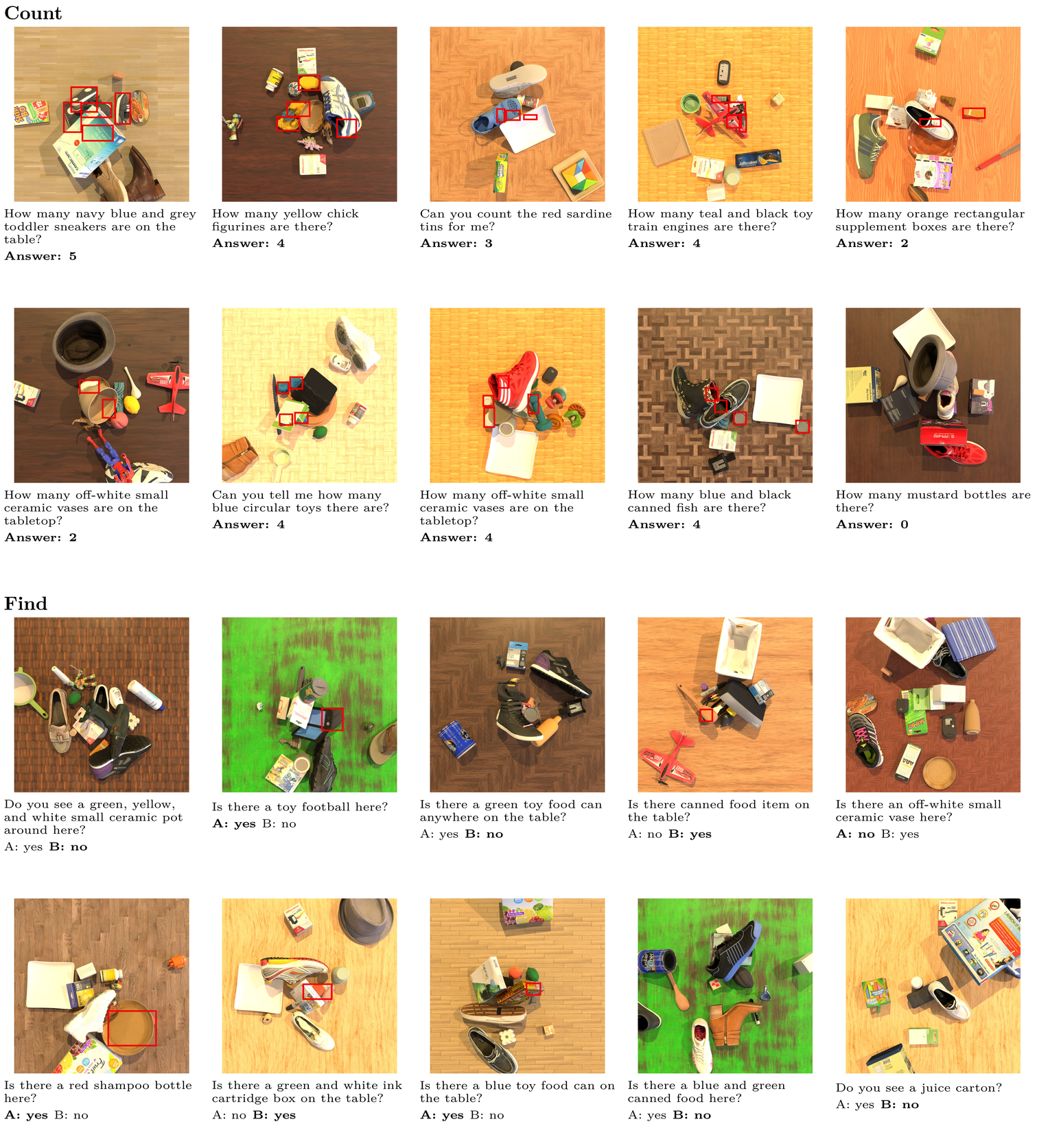}
\caption{\textbf{\projectname-Bench: Count and Find.} Ten examples per task.
Correct answers are in bold. Red boxes show ground-truth target objects;
absent targets are not boxed.}
\label{fig:bench_count_find}
\end{figure}
\clearpage

\section{Environment and Benchmark Construction}
\label{app:environment}

\subsection{Benchmark Scene Construction}
\label{app:scene_generation}

We construct candidate scenes by sampling up to 15 objects from the annotated asset pool, placing them in the tabletop workspace, and simulating settling. We use simulator state only to construct ground truth and validate benchmark instances; agents receive rendered observations and tool outputs rather than privileged object poses or labels.

\paragraph{\tCount.}
We sample a queried object category or name and a target multiplicity between 0 and 5. If targets are present, they are spawned before distractor objects, increasing the likelihood that subsequent objects occlude them after physical settling. We retain instances in which answer-critical target objects have at most 10\% pixel visibility in the initial top-down RGB observation.

\paragraph{\tFind.}
We construct equal numbers of present and absent instances. For present cases, the queried target is placed before distractors and retained only when its initial pixel visibility is at most 10\%. For absent cases, no object matching the queried name or category is present in the scene. A positive answer requires both identifying that the target is present and localizing one visible or revealed instance.

\paragraph{\tBeneath.}
After settling, we enumerate candidate cover--hidden-object pairs. For each candidate cover object, we render a counterfactual image in which that cover is transparent and measure the resulting visibility increase for nearby objects. We select pairs for which removing the cover substantially reveals a hidden object. We retain instances in which the hidden answer object has at most 10\% pixel visibility in the initial observation. For questions where the answer is ``None of the above", the hidden object is not present as any of the other 3 options.

\paragraph{\tCompare.}
We select three texture-mappable objects with distinct physical volumes. For each object, we identify a sufficiently large upward-facing mesh patch, verify that it lies in the camera frame and is self-visible in isolation, and bake one of the standardized A/B/C labels into its texture map. We construct the scene such that at least one of the labels required to associate A/B/C with their corresponding objects is initially occluded. The question asks for the largest or smallest of the three labeled objects.


\subsection{Human Verification Interface}
\label{app:human_verification}

Each candidate instance is independently reviewed by two human annotators using a custom verification interface (\cref{fig:human_verification_interface}). Annotators assess three properties: (i) whether the natural-language question, answer options, and ground-truth answer are unambiguous; (ii) whether the initial observation withholds answer-critical evidence; and (iii) whether feasible clutter removal reveals sufficient evidence to answer the question.

The interface restores the initial simulator state and displays an auxiliary verification rendering in which non-answer-critical clutter is made transparent. It highlights the evidence needed to determine the answer: target instances for \tCount{} and \tFind{}, the queried cover and hidden object for \tBeneath{}, and the objects bearing the A/B/C labels for \tCompare{}. This view is available only to annotators during benchmark construction and is never exposed to evaluated agents. For \tCompare{}, annotators additionally verified that the chosen objects are visibly distinct in occupied volume and thus can be judged for their volume comparison in the top-down view.

When annotators find that a valid scene has ambiguous wording or answer options, they edit the question metadata. We exclude instances with invalid ground truth, irrecoverable answer-critical evidence, or unresolved ambiguity.

\begin{figure*}[t]
\centering
\includegraphics[width=0.98\textwidth]{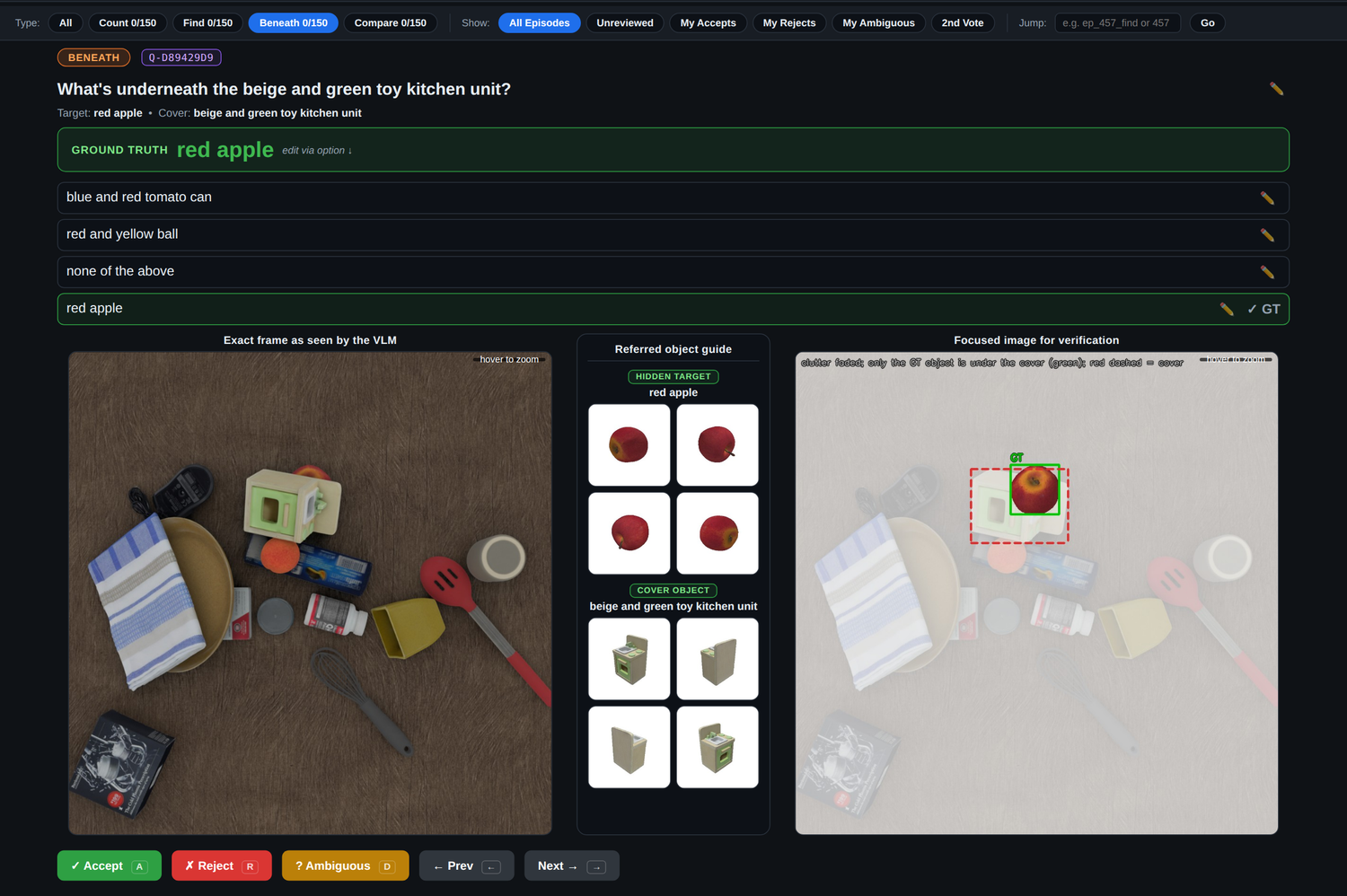}
\caption{\textbf{Human verification interface.} The custom review application displays the initial observation (left), reference object views (center), and a focused verification image (right) for the red-apple \tBeneath{} example. The privileged verification view is not exposed to evaluated agents.}
\label{fig:human_verification_interface}
\end{figure*}

\subsection{MG-VQA-Sim Implementation Details}
\label{app:implementation}\label{app:simulation_stack}\label{app:sim_comparison}

\projectname-Sim is implemented in PyBullet~\citep{coumans2021}, which simulates gravity, friction, rigid-body contact, grasping, and pushing. The environment uses a UR5 manipulator equipped with a Robotiq 2F-140 parallel-jaw gripper. We render $1024 \times 1024$ top-down RGB observations using Blender-EEVEE.

Our asset pool contains 319 real-scanned everyday objects collected from MegaPose~\citep{labbe2022megapose}, Digital Twin Catalogue~\citep{Dong_2025_CVPR}, and the Benchmark for Pose Estimation~\citep{nguyen2025bop}. For every object, we produce a short name, color, category, and physical description from multi-view renders using Gemini~3~Flash, followed by manual review. This review ensures that annotations are view-agnostic and do not depend on a particular rendered viewpoint.

Each episode contains a UR5 robot, a clearing basket within the robot workspace, and up to 15 objects placed in a $50 \times 50$\,cm tabletop area. The clearing basket allows the agent to temporarily relocate a grasped object while inspecting clutter beneath or around it. To vary visual appearance across episodes, we sample one of 108 tabletop textures from AmbientCG.\footnote{\url{https://ambientcg.com/}; CC0 1.0 Universal.} After object placement, we simulate 30 seconds of settling before rendering the initial observation. We retain scenes in which all initialized objects remain within the top-down camera field of view.

\subsection{Tool APIs}
\label{app:toolshed}

Each episode exposes four information-gathering tools and one terminal answer tool. The VLM may invoke information-gathering tools repeatedly before submitting its response through \code{answer\_question}.

\paragraph{\code{perceive\_scene}.}
This tool returns a top-down scene graph, a brief textual description of the workspace, a list of visible objects and their 2D image points, and pairwise spatial-relation predicates. Its output summarizes the currently visible state of the scene and can change after manipulation. All \code{perceive\_scene} tool calls use Gemini~3.6~Flash to generate the scene graph, regardless of the evaluated agent model.

\paragraph{\code{find\_object}.}
The agent supplies a 2D image point from the scene graph and this tool returns the continuous segmentation mask containing that point, using SAM3~\citep{carion2025sam}. This mask is returned as an explicit perception output and serves as input to the grasping pipeline.

\paragraph{\code{grasp}.}
The grasp tool acts on the segmentation mask generated by \code{find\_object}. We it along with the top-down depth map, and camera intrinsics to obtain a 3D object representation, then invoke GraspGen~\citep{murali2025graspgen} to generate a 6-DoF grasp proposal. If execution succeeds, the object is lifted and placed into the clearing basket. A grasp may fail because the predicted mask is inaccurate, no feasible grasp proposal survives filtering, the selected pose is not reachable, or contact with nearby clutter prevents a successful lift.

\paragraph{\code{push}.}
The push tool acts on the segmentation mask generated by \code{find\_object} in the previous step and one of four directions by the VLM: left, right, up, or down. It uses the mask to first compute the object's 2D centroid, 
The robot moves its closed gripper to an empty location on the opposite side of the object centroid and executes a linear sweep. Pushing provides an alternative to grasping for creating local clearance, moving broad or flat objects, and exposing partially occluded regions. Because pushes involve contact with dense clutter, they may produce partial displacement or move nearby objects in addition to the intended target.

\paragraph{\code{answer\_question}.}
This terminal tool records the model's answer and ends the episode. For \tFind{}, a positive answer additionally requires a 2D point locating a target instance.

System and user prompts used by VLMs in \projectname-Bench interactive evaluations are provided in \cref{fig:prompt-task-system} and \cref{fig:prompt-task-user} respectively.

\begin{figure*}[!ht]
\footnotesize
\setlength{\parskip}{2pt}\setlength{\parindent}{0pt}

\begin{tcolorbox}[promptbox, title={System prompt}]

\tcbsubtitle[promptsubtitle]{Role and tool access}
You are a spatial reasoning expert answering questions about cluttered tabletop
scenes. Tool access depends on the evaluation mode: \emph{Direct} (one image,
no tools), \emph{Perception} (scene perception, segmentation, answering), or
\emph{Manipulation} (the above, plus grasping and pushing).

\tcbsubtitle[promptsubtitle]{Tools}
\begin{toollist}
  \item[\tool{perceive\_scene()}] Returns a YAML scene graph of visible objects
    with descriptions, spatial relations, and object points.
  \item[\tool{find\_object(point)}] Segments one object with SAM3 at a
    scene-graph point and returns the mask.
  \item[\tool{grasp()}] Attempts to grasp the segmented object and place it in a
    clearing basket. A failure may still shift objects; follow the reported outcome.
  \item[\tool{push(direction)}] Pushes the segmented object left, right, up, or
    down by 10\,cm in image coordinates. Neighbouring objects may also move.
  \item[\tool{answer\_question(answer)}] Submits the answer and ends the episode.
\end{toollist}

\tcbsubtitle[promptsubtitle]{Interaction constraints}
Call exactly one tool per turn. Manipulation draws on a shared budget of 10
grasp or push calls; perception is free. Once the budget is exhausted, only
perception and answering remain, and five consecutive perception-only calls
terminate the episode. Answer when the evidence is sufficient; otherwise
investigate a relevant occluder.

\tcbsubtitle[promptsubtitle]{Response format and answer schemas}
Return one JSON object with no surrounding prose: \tool{answer} and
\tool{reasoning} in \emph{Direct} mode, or \tool{tool}, \tool{reasoning}, and
the tool arguments otherwise. Answers are a non-negative integer
(\tCount{}), A or B (\tFind{}), A--D (\tBeneath{}), or A--C for the largest or
smallest physical-volume host of circle-A, square-B, or triangle-C
(\tCompare{}; all badges exist, and their standardized design is not a volume
cue). A yes answer to \tFind{} additionally requires a \tool{point} on the
visible object, given as $[\mathrm{row},\mathrm{col}]$ on $[0,1000]$ with the
origin at the top-left.

\tcbsubtitle[promptsubtitle]{Example responses}
{\scriptsize\ttfamily\raggedright
\{"tool": "perceive\_scene", "reasoning": "Inspect the scene."\}\par
\{"tool": "find\_object", "point": [445, 697], "reasoning": "Segment the can."\}\par
\{"tool": "grasp", "reasoning": "Remove the segmented object to reveal what is underneath."\}\par
\{"tool": "push", "direction": "left", "reasoning": "Create space before grasping."\}\par
\{"tool": "answer\_question", "answer": "B", "point": [340, 520], "reasoning": "Target visible; B means yes."\}\par
}
\end{tcolorbox}

\vspace{-4pt}
\caption{System prompt for VLMs evaluated in \projectname-Bench, identical
across all models. Episodes terminated by five consecutive perception-only
calls or by exhausting the manipulation budget are scored as incorrect.}
\label{fig:prompt-task-system}
\end{figure*}

\begin{figure*}[!ht]
\footnotesize
\setlength{\parskip}{2pt}\setlength{\parindent}{0pt}

\begin{tcolorbox}[promptbox, title={User prompt \textnormal{(repeated at every model turn)}}]
\textbf{Status.}\quad \ph{perception-only, or remaining manipulation budget}\ \
Step: \ph{step}.\ \ Last tool: \ph{last tool}.

\textbf{Previous tool response.}\quad \ph{scene graph, segmentation result, or
reported manipulation outcome}. Omitted on the first turn.

\textbf{Question.}\quad \ph{original question}\ \
\textbf{Task.}\quad \ph{count \textbar\ find \textbar\ beneath \textbar\ compare}\ \
\textbf{Answer options.}\quad \ph{options, if applicable}

\textbf{Image.}\quad Current top-down RGB. Previous images and tool responses
remain in the conversation history. Respond using the JSON schema specified in
the system prompt.

\tcbsubtitle[promptsubtitle]{Search guidance \textnormal{(ablation only)}}
One task-specific sentence is appended, on the first turn only or on every turn
in \emph{Search-persistent}. For \emph{\tCount{}}: instances of the target may be
buried in the pile, so clear obstructing objects before committing to a number;
\emph{\tFind{}}: the target may be present but hidden, so not seeing it in the
current view is not evidence of absence; \emph{\tBeneath{}}: the answer lies
underneath the referred object, so move that object before selecting an option;
\emph{\tCompare{}}: some badges may be occluded, so uncover the ones you cannot
see before comparing.

\tcbsubtitle[promptsubtitle]{Direct mode}
A single turn containing only the question, task, answer options, and image.

\end{tcolorbox}

\vspace{-4pt}
\caption{User prompt for VLMs evaluated in \projectname-Bench, supplying the
current observation, the previous tool result, the question with its answer
options, and optional ablation context. Values substituted at runtime are shown
as \ph{placeholders}.}
\label{fig:prompt-task-user}
\end{figure*}

\section{\projectname-Hard Examples}
\label{app:hard_scenes}
We show sample question and answer pairs using
using their initial top-down observations in \cref{fig:example_hard_questions}.

\begin{figure}[!htp]
\centering
\includegraphics[width=\textwidth,height=0.78\textheight,keepaspectratio]{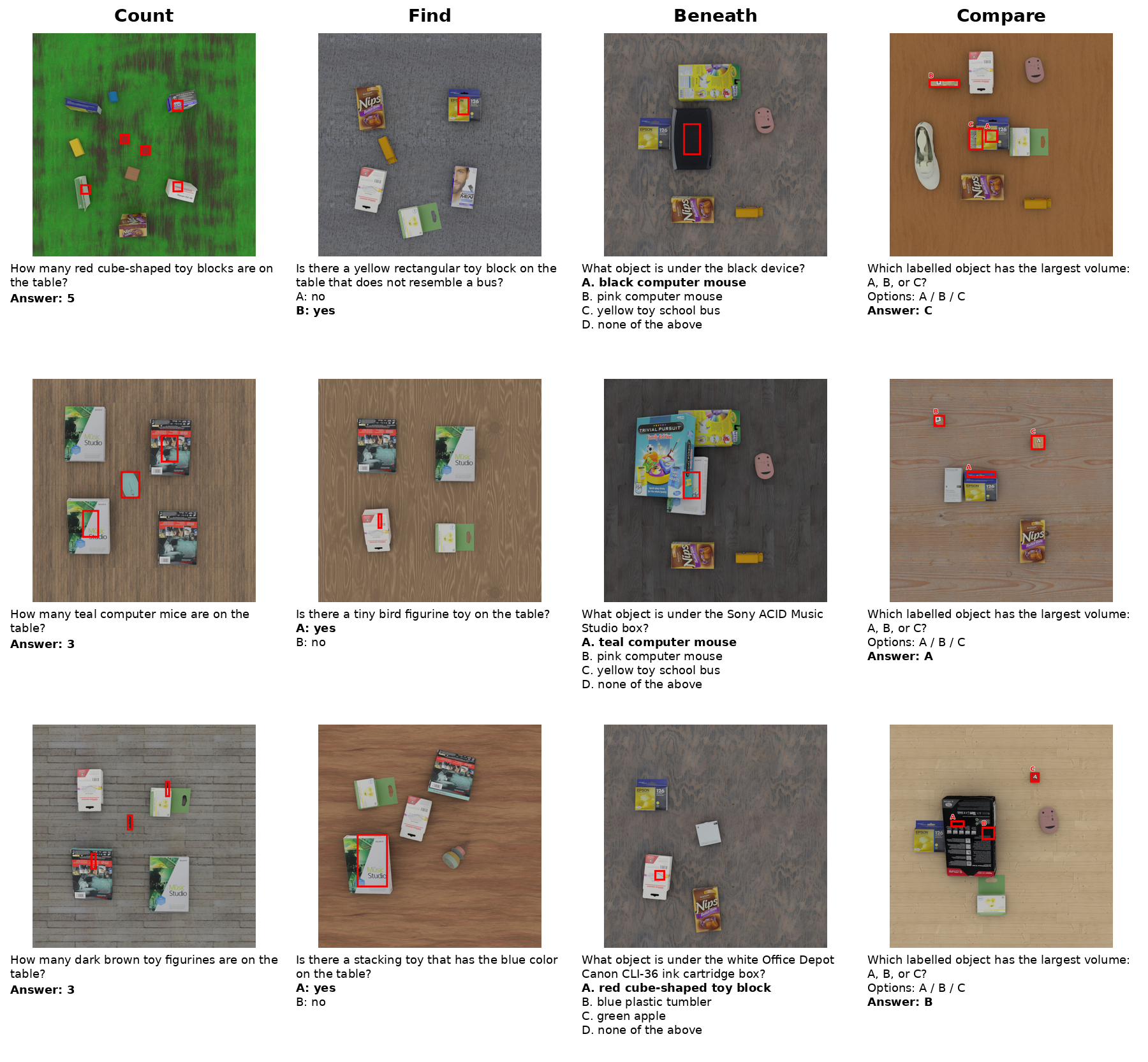}
\caption{Sample scenes in \projectname-Hard. Three per task.
Correct answers are bold. Red boxes show privileged ground-truth target
extents, including hidden portions; these overlays are not agent inputs.}
\label{fig:example_hard_questions}
\end{figure}

\clearpage

\end{document}